\documentclass[letterpaper, 10pt, conference]{ieeeconf}  

\IEEEoverridecommandlockouts                              

\usepackage{dblfloatfix}
\usepackage{graphicx} 
\usepackage{amsmath} 
\usepackage{amssymb}  
\usepackage{kantlipsum}
\usepackage{multirow}
\usepackage{cite}
\usepackage[normalem]{ulem} 
\usepackage{mathtools}
\usepackage{algorithm}
\usepackage{tabulary,booktabs}
\usepackage{subfigure}
\usepackage{algpseudocode}

\begin{document}
\title{\Large \bf Iterative Imitation Policy Improvement for Interactive Autonomous Driving}
\author{Zhao-Heng Yin$^{1}$, Chenran Li$^{2}$, Liting Sun$^2$, Masayoshi Tomizuka$^2$, Wei Zhan$^2$ 
\thanks{$^{1}$Zhao-Heng Yin is with the Department of Electronic and Computer Engineering, The Hong Kong University of Science and Technology, Hong Kong SAR. The work was conducted during his visit at University of California, Berkeley. \texttt{zhaohengyin@gmail.com}}%
\thanks{$^{2}$ Chenran Li,  Liting Sun, Masayoshi Tomizuka, and Wei Zhan are with the Department of Mechanical Engineering, University of California, Berkeley, Berkeley, CA 94720, USA. 	\texttt{\{chenran, litingsun, tomizuka, wzhan\}@berkeley.edu}}%
}

\maketitle

\begin{abstract}
We propose an imitation learning system for autonomous driving in urban traffic with interactions. We train a Behavioral Cloning~(BC) policy to imitate driving behavior collected from the real urban traffic, and apply the data aggregation algorithm to improve its performance iteratively. Applying data aggregation in this setting comes with two challenges. The first challenge is that it is expensive and dangerous to collect online rollout data in the real urban traffic. Creating similar traffic scenarios in simulator like CARLA for online rollout collection can also be difficult. Instead, we propose to create a weak simulator from the training dataset, in which all the surrounding vehicles follow the data trajectory provided by the dataset. We find that the collected online data in such a simulator can still be used to improve BC policy's performance. The second challenge is the tedious and time-consuming process of human labelling process during online rollout. To solve this problem, we use an A$^*$ planner as a pseudo-expert to provide expert-like demonstration. We validate our proposed imitation learning system in the real urban traffic scenarios. The experimental results show that our system can significantly improve the performance of baseline BC policy.

\end{abstract}
\section{Introduction}
We consider solving the autonomous driving problem in real urban traffic with imitation learning~\cite{il_survey} in this paper. The major challenge is that driving in urban traffic usually involves complicated multi-vehicle interaction in various kinds of road structures. Estimating the intention and motion of the surrounding vehicles and responding properly can be a difficult problem. The advance of supervised learning have attracted researchers to train end-to-end control policies for such problem~\cite{alvinn, chauffeur}, which directly map the observation of surrounding environment to control. Such approach is also termed as Behavioral Cloning~(BC)~\cite{il_survey}, and we also follow this approach in this paper. One benefit of BC is that can integrate perception and control into an end-to-end trainable model~\cite{cvpr2020_dagger}. Another benefit of BC is that it can be trained directly on the collected real traffic data avoiding the simulator designing issue compared with reinforcement learning based methods~\cite{rl}.
\begin{figure}[t]
    \centering
    \includegraphics[width=0.45\textwidth]{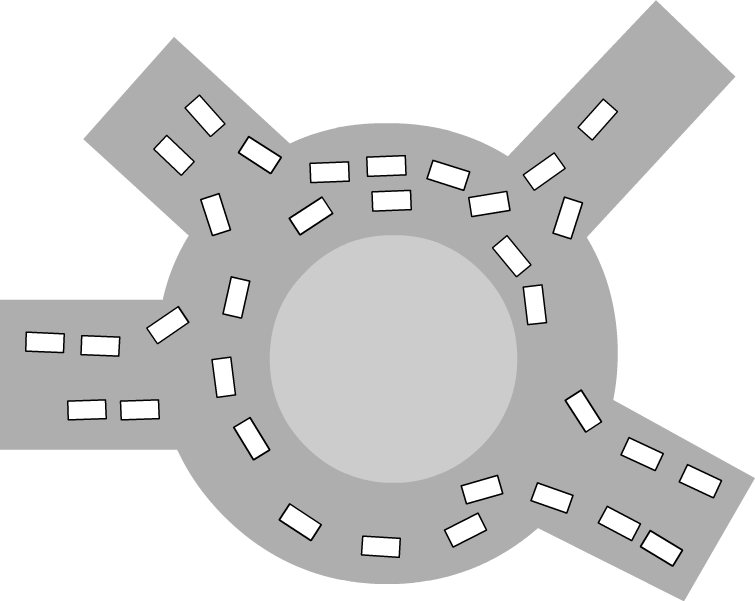}
    \caption{An illustration of urban traffic at roundabout. It is difficult to model such kind of traffic by heuristic rules for online policy rollout in DAgger. Instead of modeling it by heuristics, we directly use traffic dataset as a weak simulator. This weak simulator can be used for policy improvement.}
    \label{fig:real}
\end{figure}

Despite its success, BC policy can still be quite brittle and non-robust in practice~\cite{dagger1, dagger2}. In the autonomous driving domain, the learned policy will output improper and erroneous control when the traffic observation does not come from the training distribution. Such error can accumulate very quickly after several control steps and then lead to fatal accidents. This problem is also called the \textit{covariate shift} problem~\cite{dagger1}. To solve this problem, one solution can be using data aggregation algorithms like DAgger. In DAgger, we first rollout the trained BC policy in the environment, and then ask some human experts to provide demonstrations~(annotations) on the encountered rollout cases. These expert demonstrations are then added into the training datasets, and can be used to retrain a better policy that has less failures in evaluation. Such procedure is repeated for several iterations until the policy finally becomes robust and reaches near-expert level.

However, applying DAgger directly can be quite problematic in our setting. First, we need to collect expert data in real urban traffic during online BC policy rollout, which can be inconvenient, expensive, and dangerous~\cite{dart}. Latest works usually tackle such problem with a simulator~\cite{safe_dagger, cvpr2020_dagger}. In a simulator, one can generate numerous test cases for evaluation and expert data collection. Nevertheless, the simulated traffic in previous work are usually generated by heurstic rules~\cite{carla, torcs}, which can be tricky and even intractable if we are considering real urban traffic~(see Figure~\ref{fig:real}). Instead, we notice that a real traffic dataset itself can define a weak simulator, in which the simulated vehicles simply follow the trajectory in the dataset. We use dataset-defined simulators for online rollout in this work. Such approach can be more convenient, since collecting real traffic data can be easier than designing behavioral rules by hand. Moreover, a lot of traffic datasets are already available~\cite{Argoverse, INTERACTION}. We find this weak simulator can be useful for policy improvement, despite the fact that the vehicles inside it is non-reactive. The second problem of DAgger is that annotating the data by human experts can be tedious and time-consuming~\cite{dart}, since there are typically several millions of frames during policy rollout. Therefore, we use an pseudo-expert based on A$^*$ planner as an alternative to provide demonstration. This is possible since we are using a dataset-defined simulator, in which the future is known. We also design some critical state selection strategies to improve the sample efficiency of annotation process. We validate the implemented imitation learning system in the real traffic provided by INTERACTION dataset~\cite{INTERACTION}, which contains multiple types of road structure and interactive scenarios. We find that the performance of the BC policy can be greatly improved with our proposed system. 

Our contributions can be summarized as follows.
\begin{itemize}
    \item We implement an imitation learning system for autonomous driving in interactive urban traffic, which is based on Behavioral Cloning. We use DAgger to improve its performance iteratively. 
    \item We solve several training issues of DAgger in this setting with weak dataset-defined simulators, pseudo-expert, and critical state sampling strategies.
    \item We validate the performance of the proposed sytem on real traffic dataset, which includes various road structure and interactive scenarios collected from the real traffic. The experimental results show that our system can significantly improve the performance of baseline BC policy.
\end{itemize}

\section{Related Work}
\subsection{Behavior Cloning~(BC)} BC~\cite{il_survey} is a subcategory of imitation learning. It directly maps the observation to the action, and can be trained in an end-to-end supervised manner. The earliest use of behavior cloning in autonomous driving dates back to ALVINN~\cite{alvinn}, which uses a 3-layer neural network for lane following. Recently, the availability of large autonomous driving dataset and the emergence of deep learning~\cite{deeplearning} attract researcher to look into the power of BC and study it on more challenging domains. \cite{end2end} proposes a convolutional neural network~\cite{deeplearning} to learn the steering command. \cite{cil} proposes a conditional imitation learning framework, which is able to respond to a navigation command in addition. \cite{chauffeur} proposes some behavioral loss functions and use synthetic scenarios to improve the performance of BC. The recent trajectory prediction models~\cite{SocialLSTM, det_intentnet, det_cvpr_ff, multipath, gmm_based2} can also be viewed as instances of BC.

The main limitation of BC is that it suffers from the covariate shift problem as suggested by \cite{dagger1, dagger2}. Concretely, when executing the learnt BC policy online, the policy can gradually shift away from the expert's state distribution. As a result, the BC policy may respond improperly, which will lead to crashes. To alleviate such problem, some work has proposed noise injection~\cite{dart}, which uses action noises to make the expert demonstration cover possible states the BC policy may encounter online. Another class of work is data aggregation~(DAgger)\cite{dagger1, dagger2}, which collects expert demonstrations during online execution. Both of these methods can increase the robustness of the trained BC policy.

Besides BC, there exist some other learning-based approaches to solve the control problem in autonomous driving. Another subcategory in imitation learning is adversarial imitation learning~\cite{gail}, and researchers have used algorithm like GAIL~\cite{gail} to tackle the control problem~\cite{mixgail}. Besides imitation learning, researchers also use reinforcement learning~\cite{rl}, affordance learning~\cite{deepdriving} followed by program-based control to solve this problem.

\subsection{Data Aggregation~(DAgger)} 
Data Aggregation tackles the covariate shift problem by online expert demonstration collection. In autonomous driving, \cite{Agile} proposes to use DAgger to solve the off road control problem. \cite{safe_dagger} proposes SafeDAgger to solve the racing problem in the TORCS racing simulator~\cite{torcs}. Recently, \cite{cvpr2020_dagger} uses DAgger to solve urban driving problem in the CARLA simulator~\cite{carla}.
In order to use DAgger for policy improvement, a simulator is usually required~\cite{safe_dagger, cvpr2020_dagger} since collecting data in reality can be expensive and dangerous, though some research still manages to collect rollout data in reality~\cite{Agile}. Defining complex traffic scenarios by heurstic rules for simulation in TORCS and CARLA as previous work can be difficult and tricky. Also, the defined traffic may not cover all the possible circumstance in reality, and can be unnatural as well. In this work, we find that a real traffic dataset can actually define a useful weak simulator for DAgger. Another problem of DAgger is that we require human experts to collect demonstration. This issue is usually solved by defining heuristic control rules or apply planning~\cite{safe_dagger, cvpr2020_dagger}. Recent work~\cite{safe_dagger, cvpr2020_dagger} also report the sampling inefficiency issue of DAgger, and define rules to select critical states for expert demonstration collection.

\subsection{Planning} Soliciting human expert to provide demonstration can be quite inefficient. Therefore in the application, some pseudo experts are constructed to provide such demonstration. Some work is able to use heuristic low level control rules as the expert in some simple domains~\cite{safe_dagger}. But generated demonstration in more complicated domain which is required to have desired properties like smoothness, mild acceleration, and safety, is usually done by planning. \cite{Agile} uses Model Predictive Control as experts. A$^*$ planning~\cite{AI} is a generic planning framework for autonomous driving and is used as expert by some work~\cite{cvpr2020_dagger}. We also use A$^*$ planner in this work. One can refer to~\cite{planning} for more knowledge about the planners in autonomous driving.
\section{Dataset-Based Simulator}
\subsection{Simulation Design}
In this paper, we run the BC policy in the real traffic data provided by the INTERACTION dataset. We design simulation~(evaluation) cases based on this dataset as follow. We first select one frame as the initialization frame, in which we randomly choose a vehicle as the ego vehicle controlled by the BC policy. Then, we launch the simulation from this selected frame. The ego vehicle will follow the control calculated by the BC policy, while all the other vehicles will simply follow the trajectory provided by the dataset. In other words, the other vehicles are non-reactive. Please note that using non-reactive vehicles can be more challenging since their static response may result in critical interaction with the evaluated ego vehicle, which increases the difficulty of control for the ego vehicle. The input of BC policy is based on physical information of nearby vehicles and road, rather than RGB camera image. Note that one can still synthesis such camera image from such physical information during simulation via rendering if required. We can only rollout the BC policy and generate additional expert data on the training dataset during the training phase as the test dataset is not accessible.

\subsection{Assumptions of the Dataset}
It is worthwhile to emphasize some crucial (implicit) assumptions on the used dataset. Firstly, we assume that we can acquire the future motion of the surrounding vehicles at any time during evaluation \textit{in the training phase}. This is possible since the surrounding vehicles are non-reactive and will simply follow the trajectories in the training dataset. Such fact is important since we require such information to avoid collision when we calculate the pseudo-expert trajectory during labeling. This assumption is actually requiring that the training dataset should be continuously recorded in the traffic. Secondly, we assume that the used dataset provides some reference paths~(routes), based on which the proposed pseudo-expert can generate the expert trajectories for retraining. These reference paths are similar to the routes provided by a navigation software, and one can design them quickly. 

%
\section{Pseudo-Expert For Imitation}
\begin{figure}[t]
    \centering
    \includegraphics[width=0.5\textwidth]{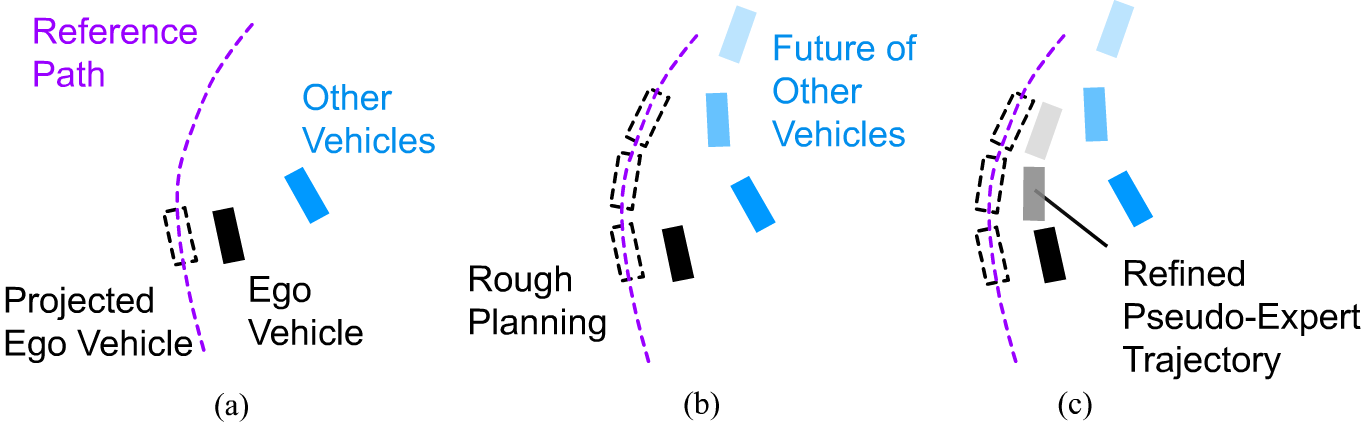}
    \caption{Illustration of the pseudo-expert trajectory generation process in an interactive scenario. (a) The ego vehicle is projected to a nearby reference path provided by the dataset. (b) We carry out planning on the reference path and get a rough pseudo-expert trajectory~(dashed black boxes). This trajectory is optimized in order to avoid collision with the other vehicles~(blue box). (c) Finally, we refined the rough trajectory and get the desired pseudo-expert trajectory.}
    \label{fig:overview}
\end{figure}
In this part, we discuss the implementation of the expert in the DAgger loop. Since asking a human expert to provide expert trajectory can be quite time consuming and tedious in practice, we implement a pseudo-expert to generate pseudo-expert trajectory for imitation. This is possible since we can know the motion of all the other vehicles in advance, which simply follow the trajectory in the dataset. 
\subsection{Overview}
We first provide an overview of the proposed pseudo-expert trajectory generation process, which is illustrated in the Figure~\ref{fig:overview}. We first project the position of the ego vehicle to the closest point on the nearest reference path~(Figure~\ref{fig:overview}a), which are provided by the training dataset. Then, we can carry out A$^*$ planning on this reference path to obtain a rough pseudo-expert trajectory~(Figure~\ref{fig:overview}b). Since this rough trajectory does not start from the current position of the ego vehicle, we apply a translation procedure to it to generate the final expert trajectory for retraining~(Figure~\ref{fig:overview}c).

\subsection{A$^*$ Planning}
A node $n_i$ in the search space is defined as a tuple $(s_i, v_i, t_i)$. Here, $s_i$ is the coordinate of the ego vehicle on the reference path, and we assume that the reference path is parameterized by curve length. $v_i$ is the velocity of the ego vehicle along the curve, and $t_i$ is the time. The beginning node is then $(s^*, v^*, 0)$, where $s^*$ is the coordinate of the projected ego vehicle on the reference path, and $v^*$ is the ego vehicle's velocity at the beginning. The transition between nodes is induced by the acceleration $a$ along the reference path. The next node of $n_i$ after taking acceleration $a$, denoted as $n_{i+1}$, is then $(s_{i+1}, v_{i+1}, t_{i+1}) = (s_i + v_i\delta t + 0.5a\delta t^2, v_i + a\delta t, t_i+\delta t)$. Here $\delta t$ is the minimum time interval. Since $a$ is a real number and can take infinite possible values, we use discretization and only allow $a\in\{-4, -3, -2, -1, -0.5, 0, 0.5, 1, 2\}$. Acceleration discretization covers common driving behaviors and is also used by previous work~\cite{acc_driving}. 
The cost function of the transition is defined as 
\begin{align*}
    \mathcal{C}&(n_{i}, a, n_{i+1}) = \\&w_1a^2 + w_2\kappa(s_{i+1}) v_{i+1}^2 + w_3(v_{i+1} - v_g)^2 + {\rm Col} (n_{i+1}).
\end{align*}
The first term penalize large tangent acceleration along the reference path. The $\kappa(s_{i+1})$ in the second term is the curvature of reference path at $s_{i+1}$. Therefore, the second term penalize large normal acceleration. The $v_g$ in the third term is a desired driving speed, which is set by prior. The third term then encourages the ego vehicle to proceed with appropriate speed. The last term ${\rm Col}(n_{i+1})$ calculates whether state $n_{i+1}$ will result in collision with the other vehicles. This function is $+\infty$ if collision will occur, and $0$ if no collision will occur. Note that calculating collision is tractable here, since other surrounding vehicles simply follow their trajectories provided by the training dataset. The planning terminates when the time of the searched node reaches the planning horizon, and the result path can be transformed to a rough expert trajectory in the world frame. In order to speed up planning process, we also discretize the search space by dividing it into uniform units whose sizes are $(\delta s, \delta v, \delta t)$. Nodes in the same unit will be identified as the same. 

\subsection{Post Planning Refinement}
Since the ego vehicle is projected onto the reference path for planning, we then translate the rough trajectory back so that this translated trajectory starts at the current position of the ego vehicle. We refine each waypoint of the rough trajectory by moving it along its normal direction, and ensure that the refined trajectory will gradually come close to the reference path. One example is illustrated in the Figure. We require that the proceeding direction~(blue arrow) on the refined trajectory should differ from that on the rough trajectory~(red arrow) by at most 20 degrees, until the ego vehicle returns to the reference path. An underlying concern for such refinement is that such refinement may lead to collision. In the practice, we find that the ego vehicle is quite close to the reference path in the majority of evaluation cases, so such refinement will still be safe. 
\begin{figure}[H]
    \centering
    \includegraphics[width=0.4\textwidth]{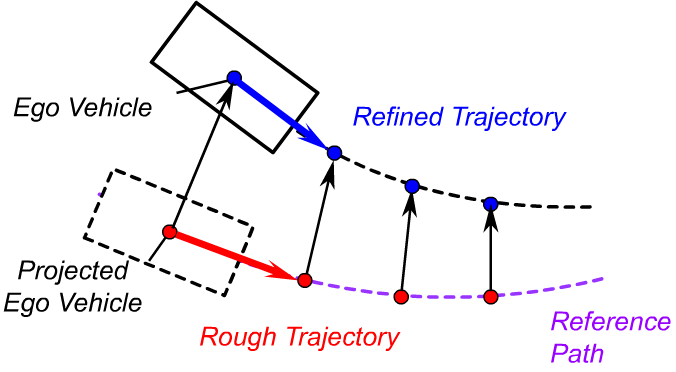}
    \caption{The illustration of refinement.}
    \label{fig:interpolate}
\end{figure}

\section{Iterative Policy Improvement}
\subsection{Critical Case Selection}
\label{section:routegan}
In the DAgger algorithm, the expert demonstrations are collected on all the encountered cases during evaluation. However, this is sample inefficient since there are usually over $5\times 10^5$ frames during the online simulation, and running the proposed pseudo-expert on all these frames will consume weeks of time. Also, it is not necessary to call the expert on a frame when the tested policy can respond properly. Hence, we should call the expert only on part of the valuable states. We design the following strategy to decide on which frame we should ask the expert for annotation.

\subsubsection{Strategy 1. $k-$Failure~(k-F.)}
The strategy 1 is quite simple and straight-forward. We will only call an expert when the evaluation fails~(crashes). Upon failure, we will call the expert starting from $k$ frames before the failure frame. We call this as $k-$Failure strategy.

\subsubsection{Strategy 2. Adversary~(Adv.)}
However, we find that strategy 1 may not be optimal in some cases, when the failure occurs due to a distant error. Also, the policy may still response improperly in the successful simulation. To percept these errors, we adopt the idea from the adversarial imitation learning. We train an discriminator to distinguish the BC policy's behavior from the expert's behavior. Then, we will use this discriminator to calculate whether the policy's behavior is like an expert in the form of probability. We will calculate such probability on all the frames in each rollout, and call the expert on the state of the lowest expert-like probability. We call this as Adversary Strategy. In practice, we find that Strategy 1 is generally enough for improvement, though the performance on some complex scenarios can be further boosted with the Strategy 2. 
\subsection{Policy Implementation}
Finally, we briefly introduce the design of the BC policy and the used low level controller in this paper. 
\subsubsection{Network Input and Output} The observation fed to the policy is an ego-centric birdeye view image of the surrounding traffic. The center of the image corresponds to the ego vehicle controlled by the policy, and the direction of the ego vehicle points upwards in the image. The image has multiple channels, including the mask of the road structure, the mask of surrounding vehicles, the velocity field of the surrounding vehicles, and the mask of the reference path. The policy also takes its current velocity as input, which is a scalar. The policy is required to output the future landmark, which will be further processed into the low level control signals. 

Note that the input observation in our experiments is different from some of the previous work, which are tested on CARLA benchmark. The observation of CARLA simulation environment is based on a simulated vehicle camera, which contains information that is irrelevant to interaction such as weather and street view. In our experiments, we focus on the policy's response to interaction and such irrelevant background information are not included in the observation. Our setting is also used by work such as the ChauffeurNet~\cite{chauffeur}.

\begin{table*}[t]
\centering
\renewcommand{\arraystretch}{1.3}
\caption{The final performance of various algorithms during online testing.}
\begin{tabular}{|l|l|l|l|l|l|l|l|l|}
\hline
\multicolumn{2}{|l|}{Algorithm}    & Scene 1 (R) & Scene 2 (R) & Scene 3 (R) & Scene 4 (I) & Scene 5 (I) & Scene 6 (I) & Scene 7 (M) \\ \hline
\multirow{3}{*}{BC}       & Suc.   & 77.4\%     & 38.8\%     & 42.5\%     & 58.4\%     & 61.2\%      & 50.8\%      & 60.1\%      \\ \cline{2-9} 
                          & Fail-V & 16.8\%     & 55.9\%     & 17.5\%     & 36.3\%     & 37.6\%      & 43.4\%      & 22.4\%      \\ \cline{2-9} 
                          & Fail-C & 5.8\%      & 5.3\%      & 40.0\%     & 5.3\%      & 1.2\%       & 5.8\%       & 17.5\%      \\ \hline
\multirow{3}{*}{A$^*$-DAgger~(k-F.)}   & Suc.   & 91.9\%     & 82.6\%     & 82.3\%     & 84.5\%     & 82.4\%      & 84.6\%      & 82.7\%      \\ \cline{2-9} 
                          & Fail-C & 6.7\%     & 16.4\%     & 16.1\%     & 13.5\%     & 16.5\%      & 15.2\%      & 15.1\%      \\ \cline{2-9} 
                          & Fail-V & 1.4\%      & 1.0\%      & 1.6\%     & 2.0\%      & 1.1\%       & 0.2\%       & 2.2\%      \\ \hline
\multirow{3}{*}{A$^*$-DAgger~(Adv.)} & Suc.   & 92.5\%     & 87.1\%     & 84.6\%     & 83.6\%     & 81.9\%      & 85.1\%      & 81.3\%      \\ \cline{2-9} 
                          & Fail-C & 6.4\%     & 12.5\%     & 14.1\%     & 13.2\%     & 16.9\%      & 14.6\%      & 16.3\%      \\ \cline{2-9} 
                          & Fail-V & 1.1\%      & 0.4\%      & 1.3\%     & 3.2\%      & 1.2\%       & 0.3\%       & 2.4\%      \\ \hline
\end{tabular}
\renewcommand{\arraystretch}{1}
\end{table*}
\subsubsection{Network structure}
We use XCeption neural network~\cite{xception} to encode the scene information. Then, the encoded features are flattened and concatenated with the current speed, and is then fed into a multi-layer perceptron~(MLP) to predict the future $n$ waypoints of the ego vehicle, which take the form of relative offset from the current position and are denoted as $[\tilde{\Delta} \textbf{x}_1, \tilde\Delta \textbf{x}_2, ..., \tilde\Delta \textbf{x}_n]$. 
\subsubsection{Loss functions}
Let the groundtruth relative offset to be $[\Delta \textbf{x}_1, \Delta \textbf{x}_2, ..., \Delta \textbf{x}_n]$. Then, the behavioral cloning loss is defined as
$$
\mathcal{L}_{pred}(\pi, \mathcal{D}) = \mathbb{E}_{\mathcal{D}}\sum_{i=1}^n\Vert\tilde\Delta \textbf{x}_i - \Delta \textbf{x}_i\Vert^2.
$$
Here, the expectation $\mathbb{E}_{\mathcal{D}}$ is taken over all the samples in the dataset $\mathcal{D}$. We also applied the following loss to avoid collision with surrounding vehicles and curbs.
$$
\mathcal{L}_{veh}(\pi, \mathcal{D}) = \frac{1}{n}\mathbb{E}_{\mathcal{D}}\sum_{i=1}^n \langle M_i, M_{veh}^i\rangle
$$
$$
\mathcal{L}_{curb}(\pi, \mathcal{D}) = \frac{1}{n}\mathbb{E}_{\mathcal{D}}\sum_{i=1}^n \langle M_i, M_{curb}\rangle
$$
In the above equations, $M_i$ is the heatmap image of the future waypoint observed by the current ego vehicle, which is calculated from $\tilde\Delta x_i$ using heatmap operation. $M_{curb}$ is the heatmap image of surrounding curbs. $M_{veh}^i$ is the heatmap image of the surrounding vehicles $i$ time steps later. $\langle\cdot,\cdot\rangle$ is the inner product of the two image (sum of elementwise product). The loss of the neural network is then defined as a weighted combination of the above loss functions:
$$\mathcal{L}(\pi, \mathcal{D}) = \mathcal{L}_{pred}(\pi, \mathcal{D}) + \lambda_1\mathcal{L}_{veh}(\pi, \mathcal{D}) + \lambda_2\mathcal{L}_{curb}(\pi, \mathcal{D})
$$
\subsubsection{Quadratic Programming refinement}
\label{section:interpolation}
Since the output waypoints calculated by the neural network may not be smooth in shape and acceleration, we also takes a quadratic programming approach to refine the waypoint prediction result. The objective is to minimize the variation of the velocity and curvature, while still keep close to the original prediction. Such refinement is used during online policy rollout. We use $\tilde{\pi}$ to denote such refinement over $\pi$. Finally, we can summarize our algorithm as below.

\begin{algorithm}[H]
    \caption{A$^*$-DAgger}
    \label{alg:overall}
    \begin{algorithmic}[1]
        \State $\mathcal{D}_0\leftarrow$ Original dataset.
        \For{$i=0, 1,2,...$}
        \State $\pi_i = \arg\min_\pi \mathcal{L}(\pi, \mathcal{D}_i)$.
        \State $\tau_i\leftarrow$ Rollout trajectories on $\mathcal{D}_0$ by $\tilde\pi_i$.
        \State Select critical states $\{s\}$ on $\tau_i$ using strategy 1 or 2.
        \State Generate new dataset $\mathcal{D}_{new}\leftarrow\{(s, \pi_{astar}(s))\}$.
        \State $\mathcal{D}_{i+1} = \mathcal{D}_{i} \cup \mathcal{D}_{new}$.
        \State [Strategy 2] Train a discriminator to distinguish samples in $\mathcal{D}_0$ from that in $\tau_i$.
        \EndFor
    \end{algorithmic}
\end{algorithm}

\section{Experiments}
\subsection{Setting}
We test our algorithm on the INTERACTION dataset. The tested scenarios include various types of road structures include Roundabout~(R), Intersection~(I), and Merging~(M). For each scenario, we split out $30\%$ cases for online evaluation.

We record the following metrics for evaluation. Fail-V metric is the collision rate with the surrounding vehicles. Fail-C metric is the collision rate with the curb. Suc. is the success rate.
\begin{figure*}[ht]
	\centering
	\subfigure[Scene 1]{
		\includegraphics[width=0.3\linewidth]{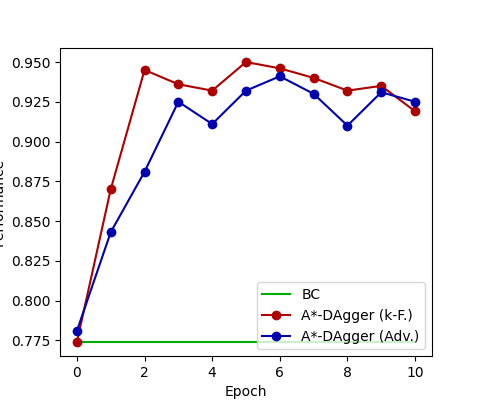}
	}
	\quad
	\subfigure[Scene 2]{
		\includegraphics[width=0.3\linewidth]{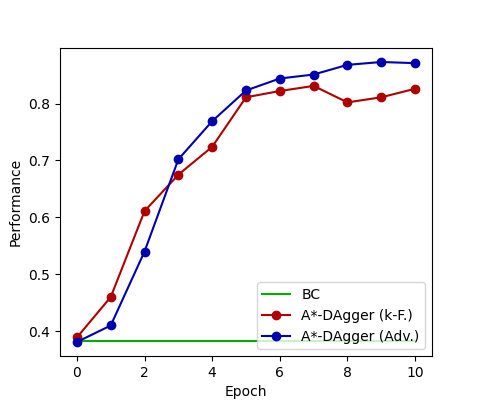}
	}
	\quad
	\subfigure[Scene 3]{
		\includegraphics[width=0.3\linewidth]{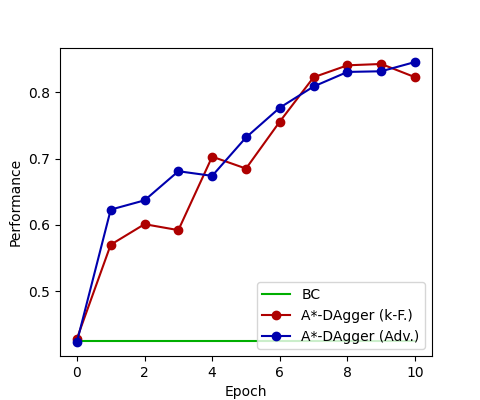}
	}
	\\
	\subfigure[Scene 4]{
		\includegraphics[width=0.3\linewidth]{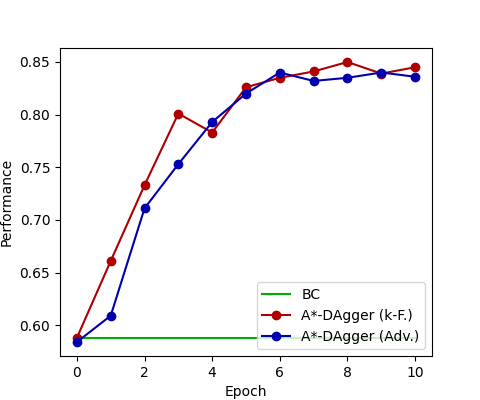}
	}
	\quad
	\subfigure[Scene 6]{
		\includegraphics[width=0.3\linewidth]{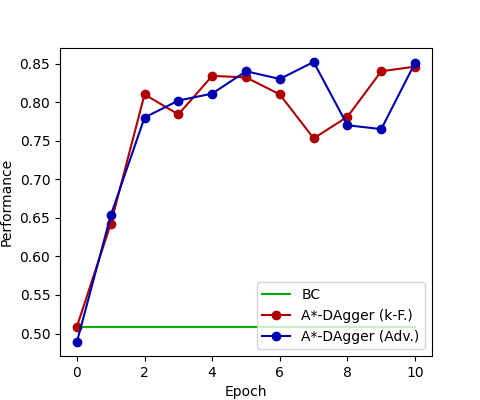}
	}
	\quad
	\subfigure[Scene 7]{
		\includegraphics[width=0.3\linewidth]{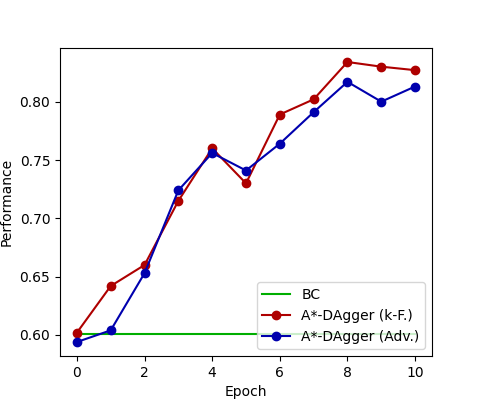}
	}
	\caption{The success rate of different algorithms after each policy improvement iteration.}
	\label{fig:curves}
\end{figure*}
\subsection{Implementation details}
Since the size of the generated dataset is much more smaller than that of the original dataset, we will duplicate the generated dataset for $10$ times before aggregating them into the latest dataset in order to balance their importance. We run our algorithm for $10$ iterations to convergence, when the performance of the policy converges on the training dataset. We also find that when the training converges, the total size of all the generated dataset is roughly the same as that of the original dataset. Rather than training policies for each scenarios seperately, we train one policy for all the scenarios jointly as we find that sharing interactions on different scenarios can lead to better generalization.

\subsection{Results}
We display the results in the Table 1. We observe that BC can only succeed on about half the test cases. In particular, it performs worse on the roundabout scenarios than on the intersection scenarios, since the roundabout scenarios involves more intensive interactions with nearby vehicles such as merging and exiting. This is different from the cases in intersection scenarios where the typical interaction mode is car following. Despite the difference between different scenarios, both of the proposed A$^*$-DAgger version can improve the original BC baseline by a large margin stably across all of these scenarios. We find that the collision with the curbs can be reduced to near zero on these scenarios. However, there still exists some collision with the other vehicles. We further find that this is partly due to the non-reactive nature of the surrounding vehicles in the simulation. For example, one failure mode is that the ego vehicle is hit by the data vehicle coming from behind. This can be a little hard to avoid since this may require the ego vehicle to accelerate and deviate from the common reference path, which is a rare pattern in the dataset. Comparing $k$-Failure strategy with Adversary strategy, we also find that A$^*$-DAgger using Adversary strategy can make a slightly improvement on roundabout. In general, two sampling strategy achieve similar performance.

\subsection{Variation of Success Rate during Training}
We also show the change of success rate on different test scenarios after each training iteration in the Figure~\ref{fig:curves}. We observe that the success rate in general increases on the first 5 iterations. However, starting from the 5th iteration, such increase gradually slows and the performance even start to drop on some scenarios a little. This is somehow due to the overfit problem of the DAgger algorithm, since the aggregated dataset may affect the original data distribution. This phenomenon is also discovered by some recent work. Our paper does not address such this issue here, and how to solve it is still an open problem.

\subsection{Qualitative Results}
\begin{figure}[t]
    \centering
    \includegraphics[width=0.485\textwidth]{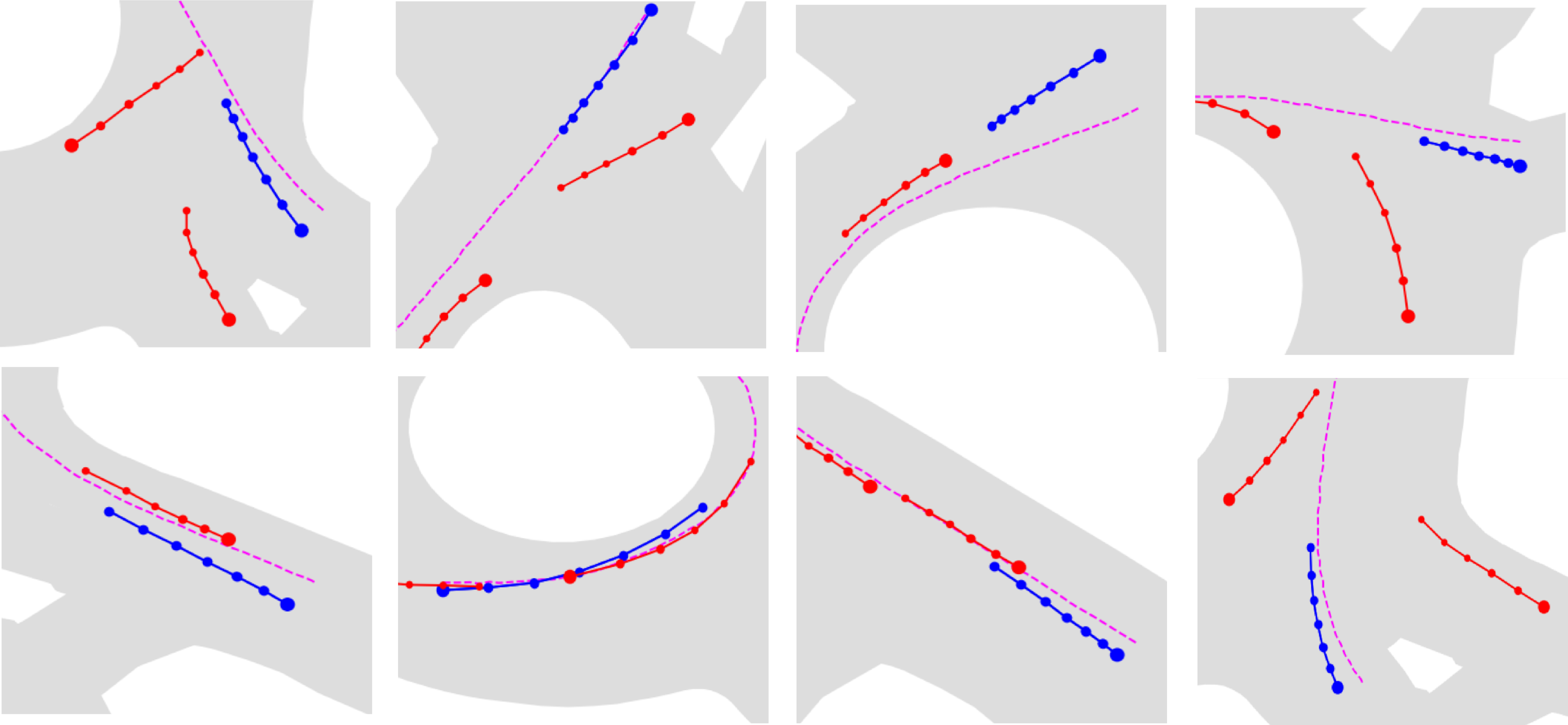}
    \caption{The visualization of the generated pseudo-expert trajectories in the roundabout scenario.}
    \label{fig:example}
\end{figure}

Finally, we visualize some generated pseudo-expert trajectory on some interactive scenarios. The results are shown in the Figure~\ref{fig:example}, which are taken from the Roundabout scenarios. In each figure, the grey background displays the road structure. The blue line indicates the generated pseudo-expert trajectory.  The other red lines indicate the motion of the other nearby vehicles. The largest dot on each solid line indicates the initial position of that vehicle. The pink dashed line indicates the reference path used by the A$^*$ planner.

We find that the pseudo expert can respond correctly on all these common cases. In particular, when entering the roundabout, it can take the motion of vehicles in the roundabout into consideration, and then slows down to wait them to pass. Such generated data can help the trained policy to acquire a much safer behavior and avoid collision. Moreover, we find that these generated trajectories are smooth and do not change rapidly in velocity, which is also up to our expectation.

\section{Conclusion}
In this work, we implement an imitation learning system for autonomous driving in interactive urban traffic, which can improve its performance iteratively by DAgger. We address several issues of DAgger in this challenging setting. Concretely, we design a weak dataset-based simulator to solve the online rollout problem. We design an A$^*$ planning algorithm as an alternative of human expert to provide demonstration. The strategy of critical selection is also investigated. We run evaluation experiment on real traffic datasets and show that our proposed algorithm can improve the performance of baseline behavior cloning agents and achieve high success rate. There can be many interesting directions for investigation in the future, such as learning to generate interactive urban traffic for such policy improvement.  
\bibliographystyle{IEEEtran}
\bibliography{references}
\end{document}